\newcommand{\R}{\mathbb{R}}
\newcommand{\C}{\mathbb{C}}
\newcommand{\N}{\mathbb{N}}
\documentclass[letterpaper, 10 pt, conference]{ieeeconf}  

\IEEEoverridecommandlockouts                              

\usepackage{graphics} 
\usepackage{graphicx}
\usepackage{amsmath} 
\usepackage{amssymb}  
\usepackage{physics}
\usepackage{float}
\usepackage[short]{optidef}
\usepackage{comment}
\usepackage{hyperref}
\usepackage{balance}
\usepackage{xcolor}
\usepackage[final]{changes}
\usepackage{bm}
\title{\LARGE \bf
Fast and In Sync: Periodic Swarm Patterns for Quadrotors 
}

\author{Xintong Du, Carlos E. Luis, Marijan Vukosavljev, and Angela P. Schoellig
\thanks{Authors are with the Dynamic Systems Lab \href{www.dynsyslab.org}{(www.dynsyslab.org)} at
the University of Toronto Institute for Aerospace Studies (UTIAS), Canada. 
Email: xintong.du@mail.utoronto.ca, carlos.luis@mail.utoronto.ca, mario.vukosavljev@mail.utoronto.ca, schoellig@utias.utoronto.ca}
}

\begin{document}
\setlength{\parskip}{0pt}

\maketitle
\thispagestyle{empty}
\pagestyle{empty}

\begin{abstract}
This paper aims to design quadrotor swarm performances, where the swarm acts as an integrated, coordinated unit embodying moving and deforming objects. We divide the task of creating a choreography into three basic steps: designing swarm motion primitives,  transitioning between those movements, and synchronizing the motion of the drones. The result is a flexible framework for designing choreographies comprised of a wide variety of motions. The motion primitives can be intuitively designed using \added{a} few parameters, providing a rich library for choreography design. Moreover, we combine and adapt existing goal assignment and trajectory generation algorithms to maximize the smoothness of the transitions between motion primitives. Finally, we propose a correction algorithm to compensate for motion delays and synchronize the motion of the drones to a desired periodic motion pattern. The proposed methodology was validated experimentally by generating and executing choreographies on a swarm of 25 quadrotors.
\end{abstract}

\section{INTRODUCTION}
In recent years, robots have gradually found their way into the entertainment industry. Due to the advancements in both software and hardware, autonomous quadrotors, for example, have become active participants in art work and entertainment activities, showing off their exceptional agility in navigating in the three-dimensional space. Companies such as \textit{Intel} and \textit{SKYMAGIC} have delivered dazzling drone shows, where hundreds or even thousands of drones equipped with dedicated LEDs were coordinated to display enormous animations in the sky. Other companies, such as \textit{Verity Studios} and \textit{ElevenPlay}, have taken their shows to a different level, involving coordinated movements of drones with human performers, light effects, and music. 

Drone shows where visual effects dominate the audience's experience have been pushed to an unprecedented scale and flown as many as 2018 drones \cite{Intel}. However, shows that primarily rely on the effect of highly dynamic motions and fully leverage the motion capabilities of quadrotors remain at a much smaller scale. One reason is that drone swarms performing highly dynamical motions demand fast and accurate state estimates, motion planners and controllers, as well as an efficient and reliable overall flight control system architecture (e.g. with little or predictable time delays). Moreover, from a producer's perspective, the design and choreography of attractive, highly dynamic, swarm-like motion patterns for a large troupe of drones becomes more challenging due to inter-agent collision constraints and spatial constraints, as well as aerodynamic effects.
Therefore, our work aims to provide motion planning and control strategies that enable a quadrotor swarm to act as an integrated entity and display highly dynamic motion patterns, as shown in Fig. 1. 

\begin{figure}[t] 
\includegraphics[width=\linewidth]{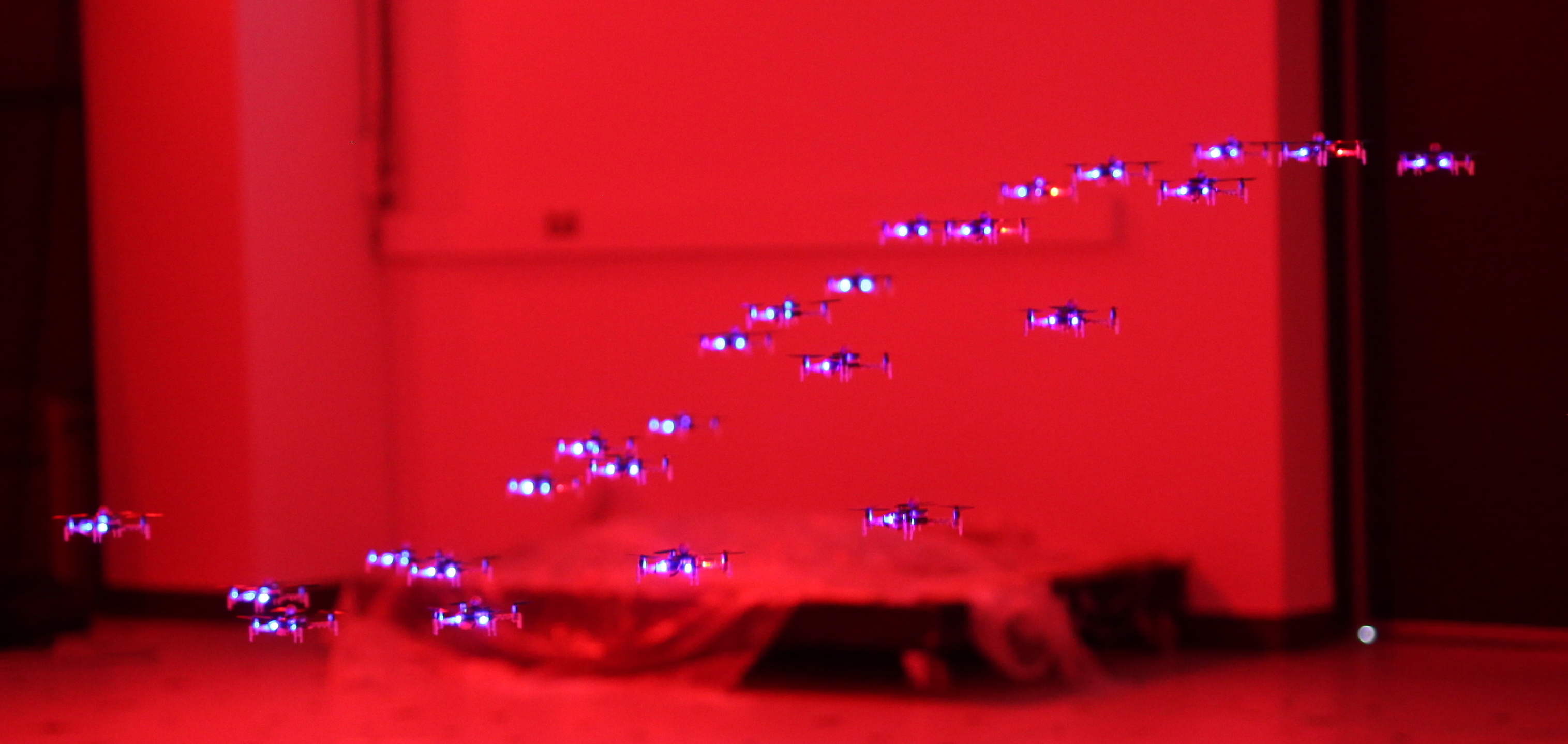}
\caption{Twenty-five quadrotors perform a periodic wave motion in the vertical direction. A video of a full performance is available at \href{http://tiny.cc/fast-periodic}{\url{http://tiny.cc/fast-periodic}}.\vspace{-0.2cm}}
\label{fig:CoverFigure}
\end{figure}

In the context of motion planning for complex task specifications, motion primitives have proven to be a successful framework. 
Motivated by creating theatrical performances, we aim to design a library of motion primitives, from which choreographers can then choose and combine motion primitives to form a coherent story. 
First, individual motion primitives must be designed.
Rhythmic motion primitives for individual vehicles were proposed in \cite{augugliaro2013methods}. A set of representative behaviour descriptors in \cite{cappo2018online} was designed to enable online interaction with a small group of robots. However, a unified framework for encoding a library of motion primitives for swarms is lacking.

Next, to concatenate swarm motion primitives, we develop an algorithm for smooth and safe transitions. Many techniques in the robot planning and control literature already exist to transition vehicles from one configuration to another. Of particular interest is a simultaneous goal assignment and trajectory planning algorithm with a simple collision avoidance scheme proposed in \cite{turpin2014goal}, which has been applied in the context of drone performances \cite{desai2016dynamically}. Despite its scalability to large swarms and its consideration of vehicle dynamics at an early stage in the planning process, this method does not immediately apply to our situation. Namely, it does not guarantee a common start and end time for all vehicles, and it assumes the vehicles are at rest at the start and end, which makes designing smooth trajectories difficult for our dynamic motions.

Finally, once motion primitives and their transitions are designed, they need to be executed reliably by the real system. To this end, a phase comparator \deleted{along with a correction algorithm} \replaced{was}{were} proposed in \cite{schollig2010synchronizing} to eliminate the phase error in tracking the desired position trajectories. \added{Moreover, a correction method was taken to compensate for the predictable delay in sensing and communication. In this work, we directly apply the correction method to quadrotors with standard position and attitude controllers. We demonstrate that this method can reliably minimize tracking errors of periodic signals for quadrotor swarms.} 

The contributions of this paper are as follows. First, we present a generic formulation for swarm motion primitives that is suitable for describing periodic, highly dynamic motion patterns, where drones appear as a coordinated unit embodying moving and deforming objects. 
Second, we adapt and combine various state-of-the-art trajectory planning methods to safely and smoothly connect arbitrary swarm motion primitives and, additionally, preserve patterns in the swarm behaviour that could emerge from strategic goal assignment. Third, we show that a simple correction algorithm is capable of compensating for amplitude and phase errors that arise when trying to synchronize a swarm's motion to a desired periodic motion pattern. Finally, results are demonstrated experimentally on a large swarm of drones.

\section{DYNAMIC, PERIODIC MOTION PATTERNS}
This section presents our design of dynamic, periodic motion primitives for drone swarms. These motion primitives embody moving and deforming objects and are inspired by natural particle phenomena such as wave motions or rigid body rotations. The goal is that vehicles appear as an integrated entity.

\subsection{Generic Formulation}
We define a motion primitive for a swarm of $N$ drones as a tuple
\begin{equation} \label{eq:swarmprim}
	\mathcal{MP}^{N} = \left(t_0, t_f, \{\mathbf{r}^n\}_{n=1}^N, \mathcal{T}_d(\mathbf{r}, t) \right),
\end{equation}
\noindent which consists of the start time $t_0 \in \R $ and end time $t_f \in \R$ of the motion primitive,
a unique characteristic configuration vector $\mathbf{r}^n \in \R^3$ for each drone $n = 1, \ldots, N$,
and a desired position trajectory generator $\mathcal{T}_d(\mathbf{r}, t) : \R^3 \times [t_0, t_f] \rightarrow \R^3$. Then, the desired position evolution for the $n$-th drone over the interval $t \in [t_0, t_f]$ is given as 
$\mathbf{x}_d^n(t) = \mathcal{T}_d(\mathbf{r}^n, t)$.

As in \cite{augugliaro2013methods}, we parameterize the trajectory generator $\mathcal{T}_d$ as a finite sum of periodic functions to encode rhythmic motions. Moreover, we use the configuration vectors $\mathbf{r}^n$ to emphasize how each drone's motion is related to the overall display. 
Specifically, trajectory function is defined as
\begin{equation} \label{eq:MPGen}
	\mathcal{T}_d(\mathbf{r}, t) = \mathcal{C}(\mathbf{r}) + 
	\sum\limits_{m=1}^M \left( \mathcal{A}_m(\mathbf{r}) \sin(\omega_m t) + \mathcal{B}_m (\mathbf{r}) \cos(\omega_m t) \right),
\end{equation}
\noindent with frequencies $\omega_m \in \R$ as well as parameter functions $\mathcal{C}$, $\mathcal{A}_m$ and $\mathcal{B}_m:  \R^3 \rightarrow  \R^3$ that define the centre and amplitude of the sinusoidal functions in 3D. Here $m$ indexes motions with different frequencies and parameter functions. 

The parameters in (\ref{eq:MPGen}) reflect desired rhythmic and visual effects. 
The temporal components $\omega_m$ describe intervals of repeated patterns.
The spatial components $\mathcal{C}$, $\mathcal{A}_m$, $\mathcal{B}_m$, and $\{\mathbf{r}^n \}$ are responsible for the spatial pattern collectively demonstrated by the swarm. In particular, the parameter functions determine the overall picture presented to the audience, where each drone's contribution is reflected by their unique characteristic configuration vector. 

Some possible choices for spatial patterns are given as below. In our experiments, we demonstrate two specific representations of (2), where vehicles represent the particle movement seen in waves and rigid body rotations.

\begin{figure}[t]
\includegraphics[width=\linewidth,trim={0.0cm 0.0cm 0.0cm 0.0cm},clip]{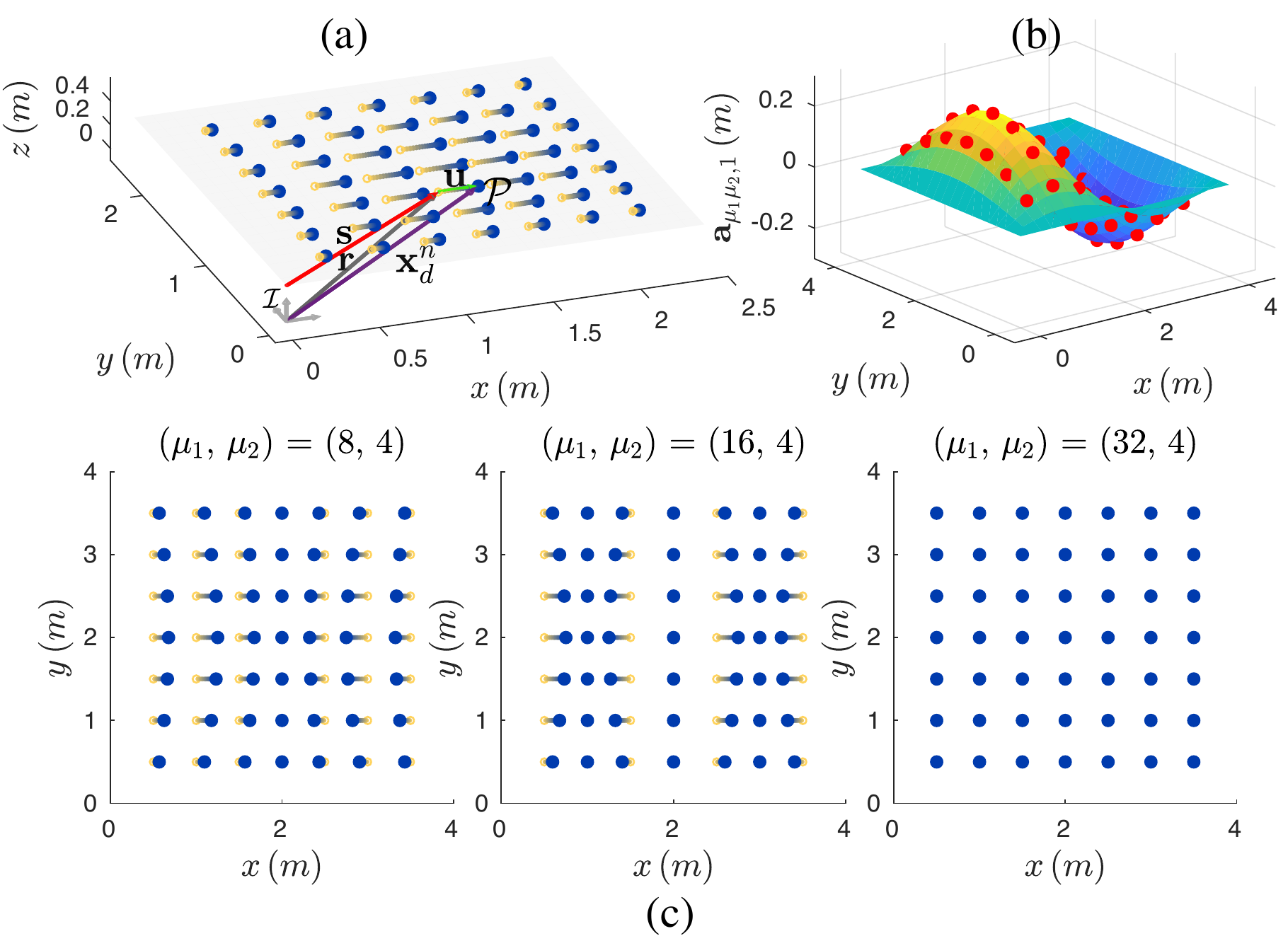}
\caption{This figure shows an example of a wave pattern on a bounded elastic surface represented by 49 drones positioned in a grid configuration. In (a), we show a schematic of a 1D wave motion over a quarter of a period where hollow circles and solid circles indicate the equilibrium and peak position, respectively. The color of the trajectories shifts from yellow to blue indicating the passage of time. In (b), the amplitude function $\mathcal{A}_m(\mathbf{r})$ in the $x$ direction is $a_{\mu_1\mu_2,1} \sin(2\pi r_1) \sin(\pi r_2)$ where $m=(\mu_1, \mu_2) = (8, 4)$. Red dots are the amplitude evaluated at each characteristic vector $\mathbf{r}^n$. The wave motion with this amplitude function is demonstrated in the bottom left diagram in this figure. In (c), we illustrate the role of the index $\mu_1$ (while keeping $\mu_2$ fixed) in determining the wave motion's periodicity and symmetry. 
\vspace{-0.2cm}}
\label{fig:2DWave}
\end{figure}

\subsection{Wave Patterns}
Wave motions are ubiquitous in nature. Typical examples include surface water waves, sound traveling through air, and electromagnetic field propagation. They can be modeled by the wave propagation equation \cite{georgi1993physics}. Consider a two-dimensional rectangular elastic surface with bounded edges as shown in Fig. \ref{fig:2DWave}a. Suppose the equilibrium location of each point $\mathcal{P}$ on the surface is parameterized by $\mathbf{s} = (s_1,s_2) \in [0,a] \times [0,b] \subset \R^2$. 
The disturbance to a point at time $t$, $\mathbf{u}(\mathbf{s},t) \in \R^3$, by a three-dimensional wave propagating through the surface at speed $c > 0$ is governed by
\begin{equation}
	c^2 \nabla^2 \mathbf{u}(\mathbf{s}, t) = \pdv[2]{\mathbf{u}(\mathbf{s}, t)}{t},
\end{equation}
where $\nabla^2$ is the spatial Laplacian. Its solution can be expressed as a product of spatial and temporal components:
\begin{equation}\label{eq:surfWave}
\begin{split}
    \mathbf{u}=
    \small{\sum_{(\mu_1,\mu_2)}}
&\mathbf{a}_{\mu_1\mu_2} \sin(\frac{\mu_1}{a}\pi s_1) \sin(\frac{\mu_2}{b}\pi s_2) 
\sin(\omega_{\mu_1\mu_2}t)+\\
&\mathbf{b}_{\mu_1\mu_2} \sin(\frac{\mu_1}{a}\pi s_1) \sin(\frac{\mu_2}{b}\pi s_2) 
\cos(\omega_{\mu_1\mu_2}t),
\end{split}
\end{equation}
\noindent where the amplitudes $\mathbf{a}_{\mu_1\mu_2},\mathbf{b}_{\mu_1\mu_2} \in \R^3$ can be determined from the surface's initial position and velocity, and frequencies $\omega_{\mu_1\mu_2} \in \R$ for $\mu_1,\mu_2 = 1,2,3 ...$ are dictated by the dispersion relation $\omega_{\mu_1\mu_2}^2 = c^2\pi^2 (\frac{\mu_1^2}{a^2}+\frac{\mu_2^2}{b^2})$.

Finally, we can use a finite approximation to this solution to generate reference trajectories for $N$ drones situated on 
\begin{figure}[t] 
\includegraphics[width=\linewidth]{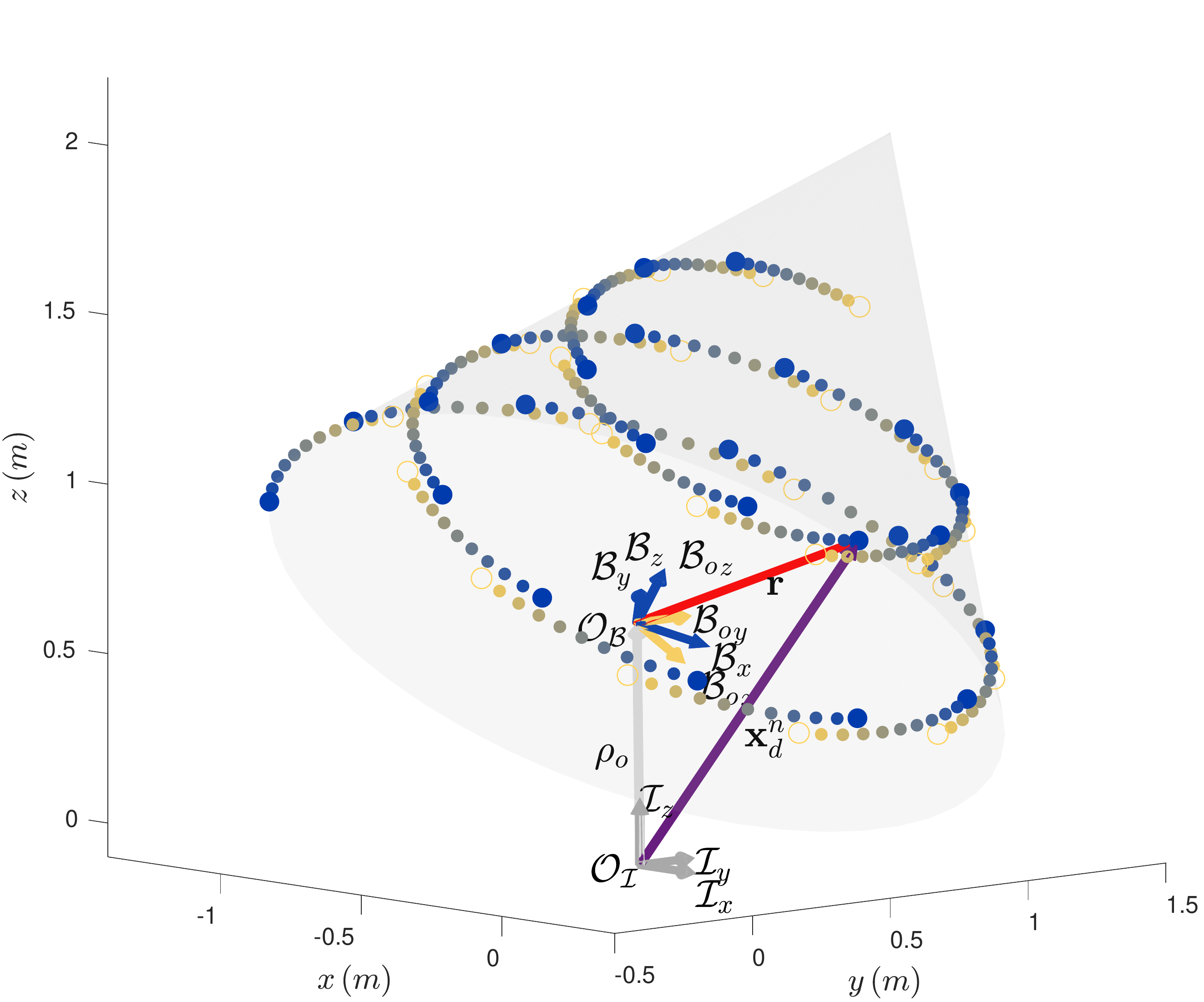}
\caption{An example of a rigid body rotation of which the characteristic vectors $\{ \mathbf{r}^n\}_{n=1}^{N}$ are designed to form a helix on the surface of a tilted cone. Desired trajectories of each drone $\mathcal{T}_d(\mathbf{r}, t)$ over a time interval $t \in[t_1, t_2]$ are shown, where hollow circles and solid circles mark the start and end position, $\mathcal{T}_d(\mathbf{r}, t_1)$,  $\mathcal{T}_d(\mathbf{r}, t_2)$, respectively. Color changes from yellow to blue along each trajectory to indicate the passage of time.\vspace{-0.5cm}}
\label{fig:Rotation}
\end{figure}
the surface. Suppose that we select $M$ pairs of $(\mu_1,\mu_2)$ terms in the finite approximation and assign a unique point $\mathbf{r}^n \in [0,a] \times [0,b] \times \{h\}$ to each drone $n = 1, \ldots, N$ on the surface at some desired height $h > 0$. Then the desired reference is $\mathbf{x}_d^n(t) = \mathbf{r}^n + \mathbf{u}((r^n_1,r^n_2),t)$ for $n = 1, \ldots, N$. 
Comparing to \deleted{\eqref{eq:swarmprim}  and} \eqref{eq:MPGen}, we can design a swarm motion primitive, where the indices $m = 1, \ldots, M$ correspond to a pair $(\mu_1, \mu_2)$, the frequencies are $\omega_m = \omega_{\mu_1\mu_2}$, and parameters are
$\mathcal{C}(\mathbf{r}) = \mathbf{r}$,
$\mathcal{A}_m(\mathbf{r})  = \mathbf{a}_{\mu_1\mu_2} \sin(\frac{\mu_1}{a} \pi r_1) \sin(\frac{\mu_2}{b}\pi r_2)$ and $\mathcal{B}_m(\mathbf{r})  = \mathbf{b}_{\mu_1\mu_2} \sin(\frac{\mu_1}{a} \pi r_1) \sin(\frac{\mu_2}{b}\pi r_2)$.

The surface wave \eqref{eq:surfWave} has desirable geometric properties, such as symmetry and periodicity in its spatial components $\mathcal{A}_m$, $\mathcal{B}_m$. However, the parameters $\mathbf{r}^n$, $\mu_1$, $\mu_2$, $\mathbf{a}_{m}$ and $\mathbf{b}_{m}$ must be carefully selected when representing a pattern on the continuous rectangular surface with a finite number of drones. To be specific, $\mu_1$ and $\mu_2$ determine the spatial frequency of the oscillation amplitude and thus, the symmetry and periodicity in the overall wave pattern. The pattern gets more interesting as the axes or points of symmetry increase until their spacing is smaller than that between drones. Although any distribution of the drones within the surface is valid and should not influence the overall pattern, a selection of the vectors $\{\mathbf{r}^n \}_{n=1}^N$ that shares similar symmetry as $\mathcal{A}_m$, $\mathcal{B}_m$ may offer a better visual experience to the audience, such as a radial or rectangular mesh.  Examples illustrating the role of each parameter are shown in Fig. \ref{fig:2DWave}b and Fig. \ref{fig:2DWave}c.

\subsection{Rigid Body Rotation}
A rigid body rotation can also be expressed in the form \eqref{eq:MPGen}. Consider a rigid body $\mathcal{V}$ rotating in an inertial frame $\mathcal{F}_\mathcal{I}$ at a constant angular velocity, for example, the tilted cone in Fig.~\ref{fig:Rotation}. First we attach a body frame $\mathcal{F}_\mathcal{B}$ with origin $\boldsymbol{\rho}_o$ relative to the inertial frame $\mathcal{F}_\mathcal{I}$. Next, we define an inertially fixed frame $\mathcal{F}_{\mathcal{B}_o}$ which overlaps with $\mathcal{F}_\mathcal{B}$ at time $t_o=0$. 

Let $\mathbf{r} \in \R^3$ be the position of a point expressed in $\mathcal{F}_\mathcal{B}$. Then this point is expressed in the inertial frame $\mathcal{F}_\mathcal{I}$ as
\begin{equation}\label{eq:RigidBodyGen}
	\mathbf{r}^{\mathcal{I}} = \boldsymbol{\rho}_o + \mathcal{R}_{\mathcal{IB}_o}\mathcal{R}_{\mathcal{B}_o\mathcal{B}} \, \mathbf{r},
\end{equation}
where $\mathcal{R}_{\mathcal{IB}_o} = [\mathbf{e}_1 \, \mathbf{e}_2 \, \mathbf{e}_3]$ is the rotation matrix from $\mathcal{F}_{\mathcal{B}_o}$ to $\mathcal{F}_\mathcal{I}$ and, without loss of generality, $\mathcal{R}_{\mathcal{B}_o\mathcal{B}}$ is a principle rotation along the $z$ axis of $\mathcal{F}_{\mathcal{B}_o}$ given by $\mathcal{R}_{\mathcal{B}_o\mathcal{B}} = \mathcal{R}_Z(\omega_z t)$. Expanding (\ref{eq:RigidBodyGen}), we obtain the trajectory generator of a rigid body rotation:
\begin{equation} \label{eq:MPRot}
\begin{split}
	\mathcal{T}_d(\mathbf{r}, t) = \boldsymbol{\rho}_o + \mathbf{e}_3 r_3 +& (\mathbf{e}_2 r_1 -\mathbf{e}_1 r_2) \sin(\omega_z t) + \\
	 &(\mathbf{e}_1 r_1 +\mathbf{e}_2 r_2) \cos(\omega_z t).
	\end{split}
\end{equation}
This may now be compared with \eqref{eq:MPGen} to obtain the parameters and frequencies. 

Since $\omega_z$ determines the periodicity and ($\boldsymbol{\rho}_o$\replaced{, $\mathcal{R}_{\mathcal{IB}_o}$) simply denotes the pose of the rotating object}{is simply the translation of the rotating object}, the characteristic vectors $\{\mathbf{r}^n\}_{n=1}^{N}$ are the only parameters to be designed. Given the availability of only $N$ drones as well as the constraints on the minimum distance among them, $\{\mathbf{r}^n\}_{n=1}^{N}$ should be strategically chosen in order to have the audience recognize the shape of the body. For instance, it is easier to identify a cone if the drones outline a helix lying on the surface of the cone, as opposed to spacing them uniformly. 



\section{TRANSITION TRAJECTORY PLANNING}

This section presents our transition trajectory planner that coordinates the swarm to make smooth and safe transitions \added{and preserves geometric features in this process}. \added{We formulate the problem in Sec. \ref{sec:TransPS} followed by our proposed method in Sec. \ref{sec:GA} and \ref{sec:TG} and justification in Sec. \ref{sec:AC}.}
\subsection{Problem Statement}{\label{sec:TransPS}}
Consider two consecutive motion primitives $\mathcal{MP}_1^N$ and $\mathcal{MP}_2^N$ defined as 
\begin{equation*}
\begin{split}
	\mathcal{MP}_1^N &= \left( t_{0,1}, t_{f,1}, \mathbf{\{r}_1^n\}_{n=1}^N, \mathcal{T}_{d,1}(\mathbf{r}, t) \right), \\
	\mathcal{MP}_2^N &= \left( t_{0,2}, t_{f,2}, \mathbf{\{r}_2^n\}_{n=1}^N, \mathcal{T}_{d,2}(\mathbf{r}, t) \right).
\end{split}
\end{equation*}

\deleted{In what follows, we assume a given common start time $t_s$ and end time $t_e$ for all drone actors,
where $ |t_s - t_{f,1}| \leqslant \epsilon_1$ and $ |t_e - t_{0,2}| \leqslant \epsilon_2$ for some small numbers $\epsilon_1$ and $\epsilon_2$.} We aim to compute:
\begin{itemize}
\item \added{A common start time $t_s$ and end time $t_e$ for all drone actors that satisfy $ |t_s - t_{f,1}| \leqslant \epsilon_1$ and $ |t_e - t_{0,2}| \leqslant \epsilon_2$ for some small numbers $\epsilon_1$ and $\epsilon_2$.}
	\item An assignment $\mathcal{M}: \{\mathbf{r}_1^n\}_{n=1}^{N} \rightarrow \{\mathbf{r}_2^n\}_{n=1}^{N}$ that assigns each drone identified by $\mathbf{r}_1^n$ in $\mathcal{MP}_1$ a unique characteristic configuration vector $\mathbf{r}_2^n$ in $\mathcal{MP}_2$. 
	\item For each drone identified as $\mathbf{r}_1^n$ in $\mathcal{MP}_1$, a smooth trajectory $\mathcal{T}_s^n(t)$ from $\mathcal{MP}_1$ to $\mathcal{MP}_2$ that respects the quadrotors' motion constraints, the flight space boundary and inter-agent collision constraints.
\end{itemize}
\added{Our preliminary results showed that it is possible to optimize $t_s$ and $t_e$ for energy consumption. In this work, however, we assume $t_s$ and $t_e$ are given and focus on the goal assignment and trajectory generation problem.}

\deleted{We address the trajectory planning problem in two steps: \textit{(i)} goal assignment and \textit{(ii)} trajectory generation. As in \cite{turpin2014goal,desai2016dynamically}, we employ a collision resolution strategy in the second step, but we additionally enforce a common start and arrival time, as well as the smoothness of the resulting collision-free trajectories to achieve smooth and coordinated transitions.}

\subsection{Goal Assignment} \label{sec:GA}
We formulated the goal assignment problem as a combinatorial optimization problem. 
In our application, the assignment $\mathcal{M}$ aims to maximize the smoothness of trajectories generated in the next step. More complicated swarm transition objectives can be defined, such as minimizing the likelihood of collisions. However, in this case, the cost function of one assignment depends on how other assignments are made, making it a challenging nonlinear optimization problem.

If we denote the cost of assigning $\mathbf{r}_1^\alpha$ to $\mathbf{r}_2^\beta$ as $\mathcal{J}_a(\mathbf{r}_1^\alpha,\mathbf{r}_2^\beta)$, the mapping $\mathcal{M}$ can be found by solving the linear assignment problem
\begin{equation}\label{eq:LinAss}
	\mathcal{M} = \underset{\mathcal{M}( \cdot )}{\mathrm{arg min}} \sum_{n=1}^{N} \mathcal{J}_a(\mathbf{r}_1^n, \mathcal{M}(\mathbf{r}_1^n))
\end{equation}
using the Hungarian algorithm \cite{kuhn1955hungarian}. To obtain the assignment cost $\mathcal{J}_a(\mathbf{r}_1^n, \mathcal{M}(\mathbf{r}_1^n))$, we solve a simplified minimum snap trajectory generation problem as in \cite{mellinger2011minimum}, for which only state continuity constraints are enforced. We then take the optimal cost function value of that problem as the assignment cost, namely,
\begin{mini}
	{\mathcal{T}_{\alpha,\beta}(\mathbf{\cdot})}{\int_{t_s }^{t_e} (\mathcal{T}{_{\alpha,\beta}}^{(4)}(\tau))^2 d\tau}
	{\label{eqn:convex}}
	{\mathcal{J}_a(\mathbf{r}_1^\alpha,\mathbf{r}_2^\beta) = }
	\addConstraint{\mathcal{T}{_{\alpha,\beta}}^{(p)}(t_s)}{ = \mathcal{T}{_{d,1}}^{(p)}(\mathbf{r}_1^\alpha, t_s)}
	\addConstraint{\mathcal{T}{_{\alpha,\beta}}^{(p)}(t_e)}{=\mathcal{T}{_{d,2}}^{(p)}(\mathbf{r}_2^\beta, t_e)},
\end{mini}
with $p=0,1,2,3,4$. Note that $\mathcal{T}{_{\alpha,\beta}(t)}: [t_s, t_e] \rightarrow \R^3$ is the transition trajectory assigning $\mathbf{r}_1^\alpha$ to $\mathbf{r}_2^\beta$ parametrized with a single polynomial curve of $P^{th}$ order in each direction
\begin{equation}\label{eq:TrajParm}
	\mathcal{T}_{\alpha, \beta}(t;\mathbf{x}_{\alpha, \beta}) = \begin{bmatrix} 
	 {\scriptstyle \sum_{p=0}^{P}} a_{p,x}t^n \, {\scriptstyle \sum_{p=0}^{P}} a_{p,y}t^n \, {\scriptstyle \sum_{p=0}^{P}} a_{p,z}t^n
	\end{bmatrix}^T,
\end{equation} where $\mathbf{x}_{\alpha, \beta} = [a_{0,x}\ldots a_{P,x} \, a_{0,y}\ldots a_{P,y} \, \,a_{0,z}\ldots a_{P,z}] \in \R^{\scriptscriptstyle 3(P+1)}$ is the coefficient vector. 

Following \cite{richter2016polynomial}, the optimization problem in \eqref{eq:LinAss} can be written in quadratic form, given as
\begin{mini}
	{\mathbf{x}_{\alpha, \beta}^n}{\hspace{0ex} \frac{1}{2} {\mathbf{x}_{\alpha, \beta}^n}^\mathbf{T} \, \mathbf{H} \, \mathbf{x}_{\alpha, \beta}^n}
	{\label{eq:LinAssQuad}}{}
	\addConstraint{\mathbf{\mathcal{H}}(\mathbf{x}_{\alpha, \beta}^n,\, \mathbf{r}_1^\alpha, \mathbf{r}_2^\beta)}{= 0},
\end{mini}
where $\mathbf{H} \in \R^{\scriptscriptstyle 3(P+1) \times 3(P+1)}$ is the hessian of the minimum snap cost function with respect to the decision variable $\mathbf{x}_{\alpha, \beta}^n$ and $\mathcal{H}$ denotes the state continuity constraints in \eqref{eq:LinAss}. Note that this problem is fast to solve, making it a suitable cost function for the linear assignment problem.
%
%
%

\subsection{Collision-Free Smooth Trajectory Generation}\label{sec:TG}
%


Given $\mathcal{MP}_1^N$, $\mathcal{MP}_2^N$ and the assignment $\mathcal{M}$, we aim to find a smooth and collision-free trajectory $\mathcal{T}_s^n$ for each drone. Inspired by \cite{chen2015decoupled}, we decouple the problem into $N$ sub-problems to avoid accounting for the collision constraints in a large joint space. However, our proposed algorithm consists of two steps: \textit{(i)} find a dynamically feasible candidate trajectory for each drone denoted as $\mathcal{T}_c^n$ and \textit{(ii)} iteratively resolve collisions in $\{\mathcal{T}_c^n\}_{n=1}^{N}$ (if any) in a sequential manner to obtain $\{\mathcal{T}_s^n\}_{n=1}^{N}$. In what follows, we parametrize both candidate and collision-free trajectory as in \eqref{eq:TrajParm} denoted as $\mathcal{T}_c^n(t;\mathbf{x}_c^n)$ and $\mathcal{T}_s^n(t;\mathbf{x}_s^n)$ respectively.

\subsubsection{Generating Candidate Trajectories} 
To generate candidate trajectories ${\lbrace \mathcal{T}_c^n\rbrace}_{n=1}^{N}$ for the assignment $\mathcal{M}$, we solve the full minimum snap problem in \cite{mellinger2011minimum} for each drone. This problem is the same as \eqref{eq:LinAssQuad} if we let $\mathbf{r}_1^\alpha = \mathbf{r}_1^n$ and $\mathbf{r}_1^\beta = \mathcal{M}(\mathbf{r}_1^n)$ but with an additional set of state constraints ${\mathbf{\mathcal{F}}_k(\mathbf{x}_c^n)}{\leqslant 0, k=1,2,\ldots,K,}$ that bound each drone's position and higher order states at each discrete time step $t_k = t_s + k(t_e - t_s)/K$. 
\deleted{with $p=0,1,2,3,4$, $n=1,2,...,N$ and $k=1,2,...,K$,} 


\vspace{0.02cm}
\begin{figure*}[t] 
\includegraphics[width=\textwidth, height=6cm]{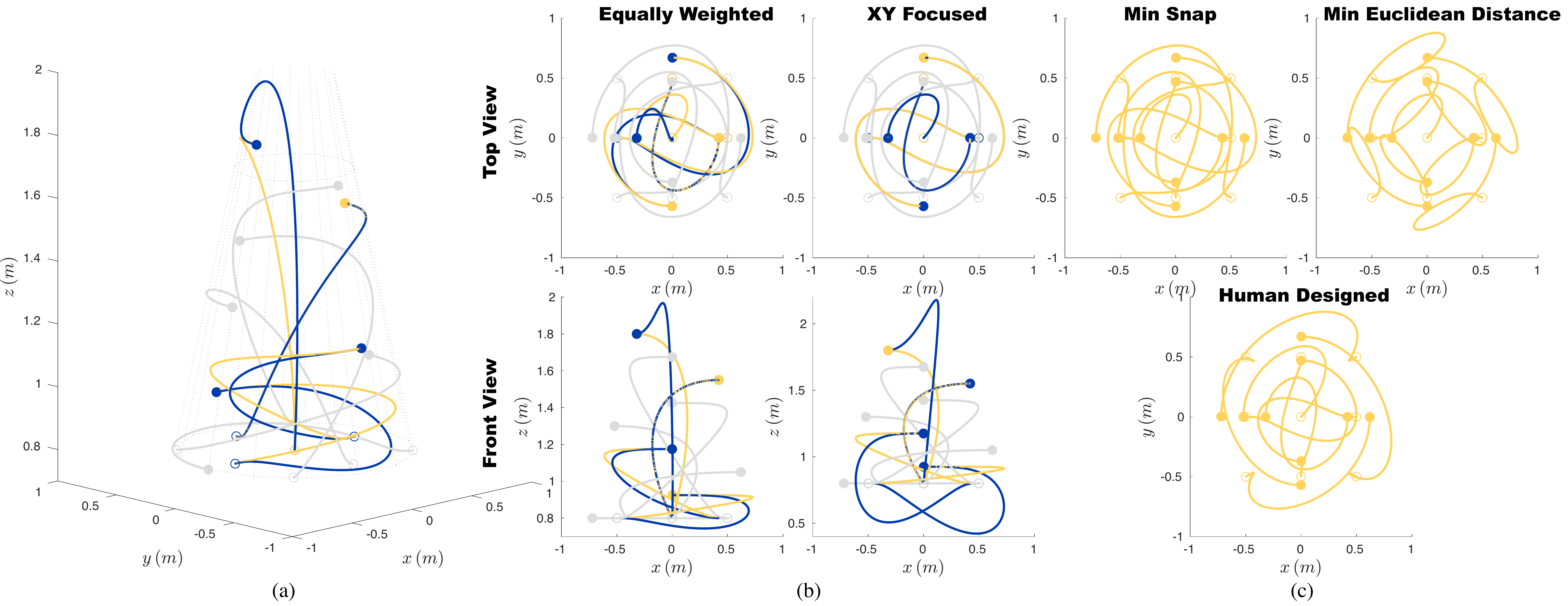}
\caption{An example of collision-free smooth trajectory generation for 9 drones from a wave pattern to a rigid body rotation similar to the ones in Fig.~\ref{fig:2DWave} and Fig.~\ref{fig:Rotation}, respectively. Hollow circles and solid circles mark the start and end of the transition trajectories, $\mathcal{T}_{d,1}(t_s)$ and $\mathcal{T}_{d,2}(t_e)$. In (a), results of the trajectory generation involving four colliding candidate trajectories $\mathcal{T}_c$ (yellow) were adapted sequentially until we obtain the collision-free $\mathcal{T}_s$ (blue). Collision-free candidate trajectories are shown in grey. In (b), we illustrate the role of weighting factor $\mathbf{W}$ in (\ref{eq:ColRes}) towards the objective of coordinated transitioning. Trajectories are colored in the same way as in (a). Finally, (c) shows comparisons between different goal assignment methods in facilitating the generation of trajectory candidates.}
\label{fig:Transition}
\end{figure*}

\subsubsection{Iterative Collision Resolution}
Given the candidate trajectories $\{\mathcal{T}_c^n\}_{n=1}^{N}$, we construct a directed collision graph $G = \{V, E\}$. A vertex $v_n \in V$ represents the drone identified by $\mathbf{r}_1^n$, and an edge $e_{nm} \in E$, $n>m$ points from $v_n$ to $v_m$, indicating a collision between $\mathcal{T}_c^n$ and $\mathcal{T}_c^m$. For any edge $e_{nm}$, we force the collision avoidance maneuver to be executed solely by drone $n$, while drone $m$ follows its intended path. Our goal is to remove all edges $e_{nm} \in E$ by finding a collision-free trajectory $\mathcal{T}_s^n$ for drone $n$.

In particular, we find $\mathcal{T}_s^n$ by penalizing its difference from $\mathcal{T}_c^n$ with additional ellipsoid collision constraints, \replaced{given as}{. Define $\mathcal{H}(\mathbf{x}_c)$ and $\mathcal{F}(\mathbf{x}_c)$ to be the set of $\widetilde{m}$ equality constraints and $\widetilde{p}$ inequality constraints in (\ref{eq:SmthTrajGen}), respectively. Then, the collision avoidance problem for drone $n$ is given as}

\begin{mini}
	{\mathbf{x}_s^n}{\hspace{3ex}\sum_{k=1}^{K} {\mathbf{e}_k^n}^\mathbf{T}\mathbf{W}\mathbf{e}_k^n}
	{\label{eq:ColRes}}{}
	\addConstraint{}{\mathcal{H}(\mathbf{x}_s^n, \mathbf{r}_1^n,\mathcal{M}(\mathbf{r}_1^n) )= 0}
	\addConstraint{ }{\mathcal{F}_k(\mathbf{x}_s^n)\leqslant 0, k=1,2,\ldots,K. }
	\addConstraint{}{||\mathbf{E}^{-1}(\mathcal{T}{_{s}^n}&(t_k) - \mathcal{T}{_{c}^m}(t_k))||_2^2\geqslant2},
\end{mini}
where $m = 1, 2, \ldots, n-1, n+1, \ldots, N$, $\mathbf{E} \in \R^{\scriptscriptstyle 3 \times 3}$ specifies the ellipsoid collision boundary as in \cite{preiss}, and ($\mathcal{H}$, $\mathcal{F}_k$) are as previously defined.
The deviation of drone $n$ at time $t_k$ is given by $\mathbf{e}_k^n = \mathcal{T}_s^n(t_k; \mathbf{x}_s^n) - \mathcal{T}_c^n(t_k; {\mathbf{x}_c^n})$. The weighting matrix $\mathbf{W} \in \R^{\scriptscriptstyle 3 \times 3}$ is a positive definite diagonal matrix trading off the deviation in each direction. Note that we can write $\mathcal{T}(t_k) = \mathbf{A}_k\mathbf{x}$ for any polynomial trajectory $\mathcal{T}(t;\mathbf{x})$ with a suitable matrix $\mathbf{A}_k$ that depends only on $t_k$. Therefore, the cost function and collision constraints in \eqref{eq:ColRes} can be written in quadratic form in $\mathbf{x}_s^n$. 

We solve (\ref{eq:ColRes}) sequentially for drones in $V$ in decreasing order of the amount of outbound edges. If an optimal solution is found for $v_n$, we remove all of its outbound edges; otherwise, in cases where the problem is temporarily infeasible or hard to find a solution (e.g., other drones are not in favourable positions), we skip to the next drone. This procedure is repeated until either $E$ is empty or the maximum iteration is exceeded. 

\subsection{Design Considerations} \label{sec:AC}
In order to generate smooth transition trajectories, a few key design decisions were made that differentiate our method from those in previous work.  First, the vehicle dynamics and transition time are incorporated into the assignment cost function $\mathcal{J}_a$ to facilitate smooth trajectory generation in the following steps. Second, trajectories are parameterized with a single polynomial curve instead of concatenated polynomials to minimize unnecessary curvature. Finally, \added{in contrast to \cite{turpin2014goal,desai2016dynamically}}, continuity in trajectories at $t_s$ and $t_e$ is guaranteed in the collision resolution step. We illustrate the first design feature in the top two figures of Fig. \ref{fig:Transition}c, which highlights the difference between two different choice of $\mathcal{J}_a$, the minimum snap function proposed in this work and the Euclidean distance used in \cite{cappo2018online}. The former assignment cost function induces smooth, fluid, and energy-saving candidate trajectories as compared to the latter. 

\added{Our choice of the cost function in \eqref{eq:ColRes} allows us to preserve geometric features in the transition process. As an example, } \replaced{the}{The} bottom figure in Fig. \ref{fig:Transition}c shows a human-designed goal assignment $\mathcal{M}$, where rotational symmetry of the initial and final motion primitives were incorporated. Consequently, it introduces nice geometric properties, for example rotation symmetry in this particular example, in the candidate trajectories $\{\mathcal{T}_c^n\}_{n=1}^N$ that enable a coordinated transition process. However, incorporating these strategic assignments into our framework is non-trivial and would result in a nonlinear assignment problem. Nonetheless, if we do have particular features in the candidate trajectories, we are able to preserve them during the collision resolution step by penalizing the difference between the final and candidate trajectories in \eqref{eq:ColRes}. Moreover, we introduced the weighting matrix $\mathbf{W}$ to optionally preserve the swarm transition patterns in some dimensions while relaxing them in the others. An example is shown in Fig. \ref{fig:Transition}b, where patterns in the $x-y$ plane are preserved during collision avoidance by encouraging motion in the $z$ direction.
 

\section{Motion Synchronization}
To execute the proposed highly-dynamic motion patterns in tight formations and in sync with the desired periodic motion pattern, a high-accuracy controller for periodic motions is required. In practice, one major issue that was observed is the phase shift and amplitude amplification or attenuation \cite{schollig2010synchronizing} due to the delay in communicating vehicle commands, the inherent delay in dynamical systems, as well as the delay in sensor measurements. Although it is possible to reduce delay and amplitude error by adding feed-forward terms to the drones' position and attitude controllers, the communication overhead it incurs is undesirable for a large swarm. Similarly to \cite{schollig2010synchronizing}, we actively adjust the desired trajectories $\mathcal{T}_d(\mathbf{r}, t)$ in (\ref{eq:MPGen}) by scaling their amplitude and shifting them in time to obtain a new reference trajectory. 

If we approximate the quadrotor as a linear system in each spatial direction with transfer functions $H_i(s)$, $i = 1,2,3$, we can estimate the amplitude attenuation and phase shift at each oscillating frequency $\mathbf{\omega}_m$ from the system's frequency response to sinusoidal reference signals. The compensated reference trajectory in the $i^{th}$ direction is given as
\begin{equation} \label{eq:Sync}
\begin{split}
	\mathcal{T}_{r,i}(\mathbf{r}, t) = \mathcal{C}_i(\mathbf{r}) 
	+\sum_{m=1}^M &\mathbf{\kappa}_{m,i} \mathcal{A}_{m,i}(\mathbf{r} ) \sin(\omega_m t+  \phi_{m,i})  \\
   +\sum_{m=1}^M &\mathbf{\kappa}_{m,i}\mathcal{B}_{m,i} (\mathbf{r}) \cos(\omega_m t + \phi_{m,i}) ,
\end{split}
\end{equation}  
where the amplitude scalings $\boldsymbol{\kappa}_m \in \R^3$ and phase shifts $\boldsymbol{\phi}_m \in \R^3$ can be determined from the closed-loop system's transfer function observed in experiments, $H_i(s)$. That is, we compute $\kappa_{m,i} = |H_i(j \omega_m)|^{-1}$, $\phi_{m,i} = - \mathrm{arg} \; H_i(j \omega_m)$. We only compensate for the motion primitives but not for the transition trajectories since the latter has an infinitely wide frequency spectrum while the former's is finite. 
\begin{figure}[t]
\includegraphics[width=\linewidth]{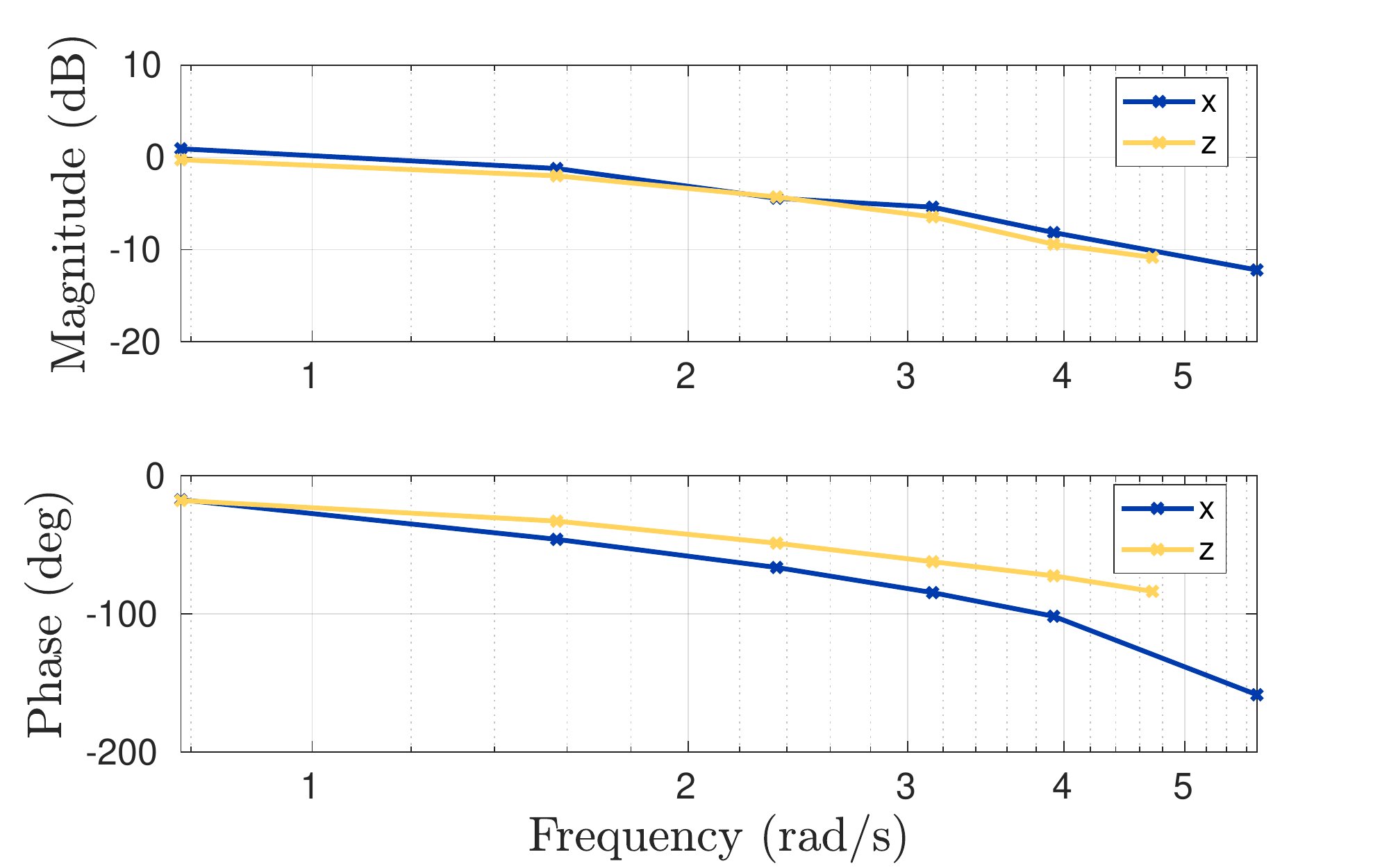}
\caption{The empirical Bode Plot for a Crazyflie 2.0 quadrotor.}
\label{fig:bode}
\end{figure}

\section{EXPERIMENTAL RESULTS}
We demonstrate our algorithms using the small-sized quadrotor platform Crazyflie 2.0 in a multi-agent testbed at the Dynamic Systems Lab (inspired by the Crazyswarm \cite{preiss2017crazyswarm}). From a central computer, we gathered position data from all the drones in the fleet using an overhead motion capture system. The state information was sent via radios to each drone's onboard computer, along with the desired position, which is tracked using an onboard position controller.


\subsection{Motion Synchronization}
\begin{figure}[t]
\includegraphics[width=\linewidth]{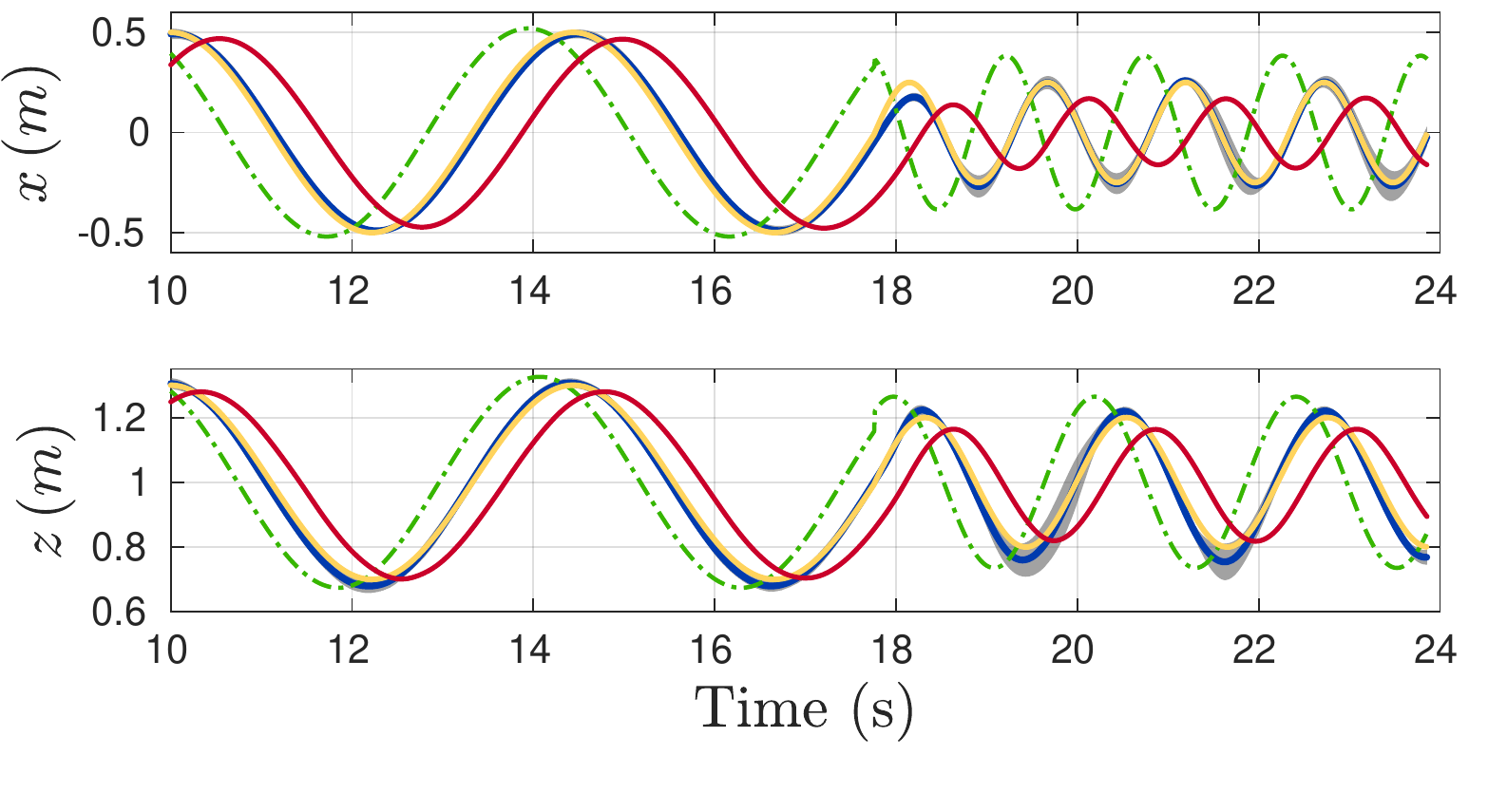}
\caption{Evaluation of the proposed motion synchronization algorithm on 8 drones. The desired trajectory (yellow) is adjusted in its amplitude and phase to obtain the reference trajectory (green). On average, the tracking response (blue) of the drones being tested is improved compared to the cases without compensation (red) with small variances (grey).}
\label{fig:comp}
\end{figure}

We empirically determined the Crazyflie's closed-loop transfer functions $H_i(j\omega)$, $i = 1,2,3$, by commanding sinusoidal position trajectories at different frequencies $\omega$ in the $x$ and $z$ directions. Based on the phase shift and amplitude attenuation between the desired and actual trajectories, we found the system's frequency response characterized by amplitude attenuation $|H_i(j{\mathbf{\omega}})|$ and phase shift $\mathrm{arg} \; H_i(j{\mathbf{\omega}})$ as shown in Fig.~\ref{fig:bode}. Based on the data, we constructed a look-up table from which the scaling factors for motion primitives, $\boldsymbol{\kappa}_m$ and $\boldsymbol{\phi}_m$, are determined by linearly interpolating points in the magnitude and phase mapping, respectively. We constructed the look-up table using data from one drone but evaluated it on another 8 drones. \replaced{As shown in Fig.~\ref{fig:comp}, the tracking performance was greatly improved with negligible variances among the 8 drones being tested}{Fig.~\ref{fig:comp} summarizes the improvements on tracking performance of the 8 drones compared to when there is no synchronization}.

\subsection{Transition Trajectory Planning}
The optimization problems in trajectory planner are solved using the nonlinear optimizer IPOPT \cite{wachter2006implementation}.
We used $P=14$ as the polynomial order and $K=10$ as a starting point for partition intervals, which may be doubled in the next iteration if a feasible solution is found but a collision occurs in between two time steps $t_k$ and $t_{k+1}$. To avoid numerical issues, we nondimensionalized the minimum snap cost function as in \cite{mellinger2011minimum}.

We evaluated the transition trajectory planner in both simulation and experiments. In simulation, we tested with 25 drones transitioning in a $ 5 {\scriptstyle \times} 5 {\scriptstyle \times} 2 \text{ m}^3$ volume with a collision radius of $ 0.14\text{m}$ in the $x$-$y$ plane and $0.35\text{m}$ in the $z$ direction. Our algorithm found a feasible solution in $85\%$ of 1800 randomly generated motion primitive pairs. The main reason for failure are numerical difficulties in solving \eqref{eq:ColRes} when the transition time is long. One repair strategy is to parameterize the trajectories with a few concatenated polynomials, instead of just one polynomial.
\subsection{Choreographed Drone Performance}
The video at \href{http://tiny.cc/fast-periodic}{\url{http://tiny.cc/fast-periodic}} presents a one-minute drone performance where 25 drones are choreographed to fly in tight formations and perform fast and dynamic motions. We seamlessly concatenate five wave motions and one rigid body rotation to create a cohesive performance. The peak velocity and pitch angle reached $2.65 \text{m/s}$ and $32$ degrees, respectively. Since the motion primitives are aggressive, we used a larger collision ellipsoid than in \cite{preiss}, with $0.28\text{m}$ in $x-y$ and $ 0.85\text{m}$ in $z$, \added{which are the radii of the smallest ellipse in $x$-$z$ plane that two drones can trace at a moderate angular velocity with $\pi$ phase shift. These parameters were coarsely estimated and worked well in our experiments. However, to fully exploit the spatial and the vehicles' physical limits, more sophisticated methods should be used to explicitly model the rotors' aerodynamics.}

%
\section{CONCLUSIONS}
In this paper, we provide guidelines for creating performances with quadrotor swarms that fully leverage their motion capabilities to create appealing visual effects. The swarm motion primitives are formulated as coupled periodic motions, which are described by a single equation indicating the overall motion pattern and the relationship between the individual actors of the swarm.
The geometric properties of parameter functions in this formulation were discussed. Moreover, we provided a hierarchical transition trajectory planner to seamlessly connect motion primitives together and preserve geometric characteristics.
Lastly, a correction algorithm is proposed to improve the quadrotors' tracking performance of the periodic motions, which allows the swarm to perform synchronously and in close proximity to each other. The method is validated with a swarm performance of 25 drones in a compact space.


\balance
\bibliographystyle{IEEEtran}
\bibliography{reference}

\begin{thebibliography}{10}
\providecommand{\url}[1]{#1}
\csname url@samestyle\endcsname
\providecommand{\newblock}{\relax}
\providecommand{\bibinfo}[2]{#2}
\providecommand{\BIBentrySTDinterwordspacing}{\spaceskip=0pt\relax}
\providecommand{\BIBentryALTinterwordstretchfactor}{4}
\providecommand{\BIBentryALTinterwordspacing}{\spaceskip=\fontdimen2\font plus
\BIBentryALTinterwordstretchfactor\fontdimen3\font minus
  \fontdimen4\font\relax}
\providecommand{\BIBforeignlanguage}[2]{{%
\expandafter\ifx\csname l@#1\endcsname\relax
\typeout{** WARNING: IEEEtran.bst: No hyphenation pattern has been}%
\typeout{** loaded for the language `#1'. Using the pattern for}%
\typeout{** the default language instead.}%
\else
\language=\csname l@#1\endcsname
\fi
#2}}
\providecommand{\BIBdecl}{\relax}
\BIBdecl

\bibitem{Intel}
Intel. (2018) Intel breaks guinness world records title for drone light shows
  in celebration of 50th anniversary. [Online]. Available:
  https://newsroom.intel.com/news/intel-breaks-guinness-world-records-title-drone-light-shows-celebration-50th-anniversary/.

\bibitem{augugliaro2013methods}
F.~Augugliaro, A.~P. Schoellig, and R.~D’Andrea, ``Methods for designing and
  executing an aerial dance choreography,'' \emph{IEEE Robot. Autom. Mag},
  vol.~20, no.~4, pp. 96--104, 2013.

\bibitem{cappo2018online}
E.~A. Cappo, A.~Desai, M.~Collins, and N.~Michael, ``Online planning for
  human--multi-robot interactive theatrical performance,'' \emph{Autonomous
  Robots}, pp. 1--16, 2018.

\bibitem{turpin2014goal}
M.~Turpin, K.~Mohta, N.~Michael, and V.~Kumar, ``Goal assignment and trajectory
  planning for large teams of interchangeable robots,'' \emph{Autonomous
  Robots}, vol.~37, no.~4, pp. 401--415, 2014.

\bibitem{desai2016dynamically}
A.~Desai, E.~A. Cappo, and N.~Michael, ``Dynamically feasible and safe shape
  transitions for teams of aerial robots,'' in \emph{IEEE/RSJ International
  Conference on Intelligent Robots and Systems (IROS)}, 2016, pp. 5489--5494.

\bibitem{schollig2010synchronizing}
A.~Sch{\"o}llig, F.~Augugliaro, S.~Lupashin, and R.~D'Andrea, ``Synchronizing
  the motion of a quadrocopter to music,'' in \emph{IEEE International
  Conference on Robotics and Automation (ICRA)}, 2010, pp. 3355--3360.

\bibitem{georgi1993physics}
H.~Georgi, \emph{The physics of waves}.\hskip 1em plus 0.5em minus 0.4em\relax
  Prentice Hall Englewood Cliffs, NJ, 1993.

\bibitem{kuhn1955hungarian}
H.~W. Kuhn, ``The hungarian method for the assignment problem,'' \emph{Naval
  research logistics quarterly}, vol.~2, no. 1-2, pp. 83--97, 1955.

\bibitem{mellinger2011minimum}
D.~Mellinger and V.~Kumar, ``Minimum snap trajectory generation and control for
  quadrotors,'' in \emph{IEEE International Conference on Robotics and
  Automation (ICRA)}, 2011, pp. 2520--2525.

\bibitem{richter2016polynomial}
C.~Richter, A.~Bry, and N.~Roy, ``Polynomial trajectory planning for aggressive
  quadrotor flight in dense indoor environments,'' in \emph{Robotics
  Research}.\hskip 1em plus 0.5em minus 0.4em\relax Springer, 2016, pp.
  649--666.

\bibitem{chen2015decoupled}
Y.~Chen, M.~Cutler, and J.~P. How, ``Decoupled multiagent path planning via
  incremental sequential convex programming,'' in \emph{IEEE International
  Conference on Robotics and Automation (ICRA)}, 2015, pp. 5954--5961.

\bibitem{preiss}
J.~A. Preiss, W.~H{\"o}nig, N.~Ayanian, and G.~S. Sukhatme, ``Downwash-aware
  trajectory planning for large quadrotor teams,'' in \emph{IEEE/RSJ
  International Conference on Intelligent Robots and Systems (IROS)}, 2017, pp.
  250--257.

\bibitem{preiss2017crazyswarm}
J.~A. Preiss, W.~Honig, G.~S. Sukhatme, and N.~Ayanian, ``Crazyswarm: A large
  nano-quadcopter swarm,'' in \emph{IEEE International Conference on Robotics
  and Automation (ICRA)}, 2017, pp. 3299--3304.

\bibitem{wachter2006implementation}
A.~W{\"a}chter and L.~T. Biegler, ``On the implementation of an interior-point
  filter line-search algorithm for large-scale nonlinear programming,''
  \emph{Mathematical programming}, vol. 106, no.~1, pp. 25--57, 2006.

\end{thebibliography}

%
%

\end{document}